\def\year{2021}\relax
\documentclass[letterpaper]{article} 
\usepackage{aaai21}  
\usepackage{times}  
\usepackage{helvet} 
\usepackage{courier}  
\usepackage[hyphens]{url}  
\usepackage{graphicx} 
\usepackage{graphicx}  
\usepackage{natbib}  
\usepackage{caption} 
\usepackage{amstext}
\usepackage[switch]{lineno}
\usepackage[section]{placeins}
\title{A Graph Reasoning Network for Multi-turn Response Selection via Customized Pre-training}

\author {
        Yongkang Liu\textsuperscript{\rm 1},
        Shi Feng\textsuperscript{\rm 1},
        Daling Wang\textsuperscript{\rm 1}\thanks{Corresponding author},
        Kaisong Song\textsuperscript{\rm 2},
        Feiliang Ren\textsuperscript{\rm 1},
        Yifei Zhang \textsuperscript{\rm 1}\\
}
\affiliations {
    \textsuperscript{\rm 1} Northeastern University, Shenyang, China \\
    \textsuperscript{\rm 2} Alibaba Group, Hangzhou, China \\
    misonsky@163.com, \{fengshi, wangdaling\}@cse.neu.edu.cn, kaisong.sks@alibaba-inc.com, renfeiliang@cse.neu.edu.cn, zhangyifei@cse.neu.edu.cn
}
\begin{document}
\maketitle

\begin{abstract}
We investigate response selection for multi-turn conversation in retrieval-based chatbots. Existing studies pay more attention to the \textbf{matching} between utterances and responses by calculating the matching score based on learned features, leading to insufficient model \textbf{reasoning} ability. In this paper, we propose a graph reasoning network (GRN) to address the problem. GRN first conducts pre-training based on ALBERT using next utterance prediction and utterance order prediction tasks specifically devised for response selection. These two customized pre-training tasks can endow our model with the ability of capturing semantical and chronological dependency between utterances. We then fine-tune the model on an integrated network with sequence reasoning and graph reasoning structures. The sequence reasoning module conducts inference based on the highly summarized context vector of utterance-response pairs from the global perspective. The graph reasoning module conducts the reasoning on the utterance-level graph neural network from the local perspective. Experiments on two conversational reasoning datasets show that our model can dramatically outperform the strong baseline methods and can achieve performance which is close to human-level.
\end{abstract}

\section{Introduction}
As an important task in a dialogue system, response selection aims to find the best matched response from a set of candidates given the context of a conversation. The retrieved responses usually have natural, fluent and diverse expressions with rich information owing to the abundant resources. Therefore, response selection has been widely used in industry and has attracted great attention in academia.

\begin{figure}[t]
\centering
\includegraphics[width=1\columnwidth]{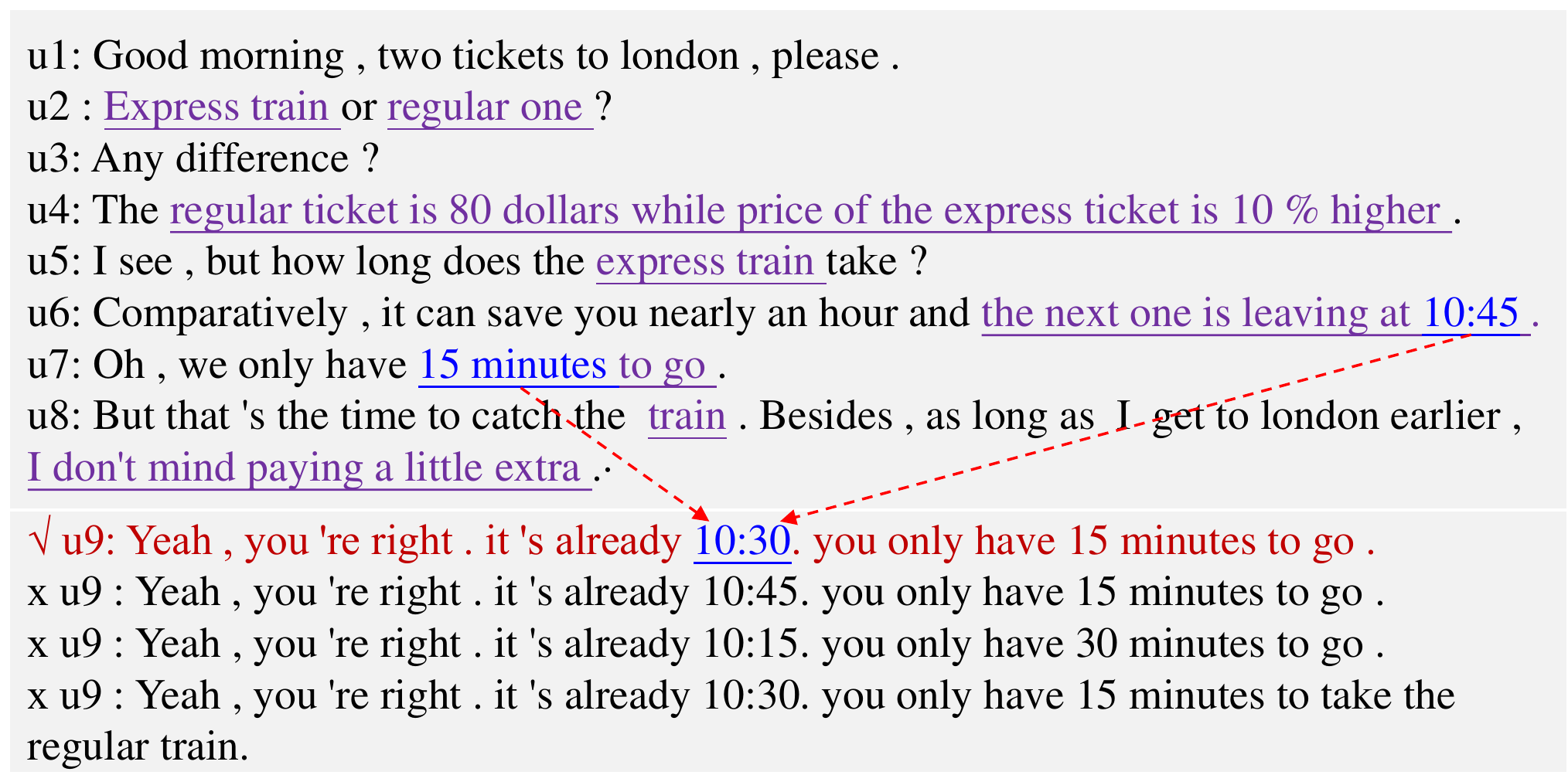}
\caption{An example from MuTual. All candidate responses are semantically relevant to utterances, but only the first one is the correct response. Clue words are purple and underlined. 
}
\label{dataem}
\end{figure}

Most existing studies on this task pay more attention to the matching problem between utterances and responses, but with insufficient concern for the reasoning issue in multi-turn response selection.
Just recently, MuTual~\cite{mutual2020}, the first human-labeled reasoning-based dataset for multi-turn dialogue, has been released to promote this line of research. \textbf{Reasoning} is quite different from \textbf{matching} in the conversations. Specifically, matching focuses on capturing the relevance features between utterances and responses, while reasoning not only needs to identify key features (i.e., words and phrases that we refer to \textit{clue words} in this paper), but also needs to conduct inference based on these clue words. The challenges of this new task include: (i) how to identify the clue words in utterances, which is fundamental for inference; (ii) how to conduct inference according to the clue words in utterances. Figure~\ref{dataem} illustrates a motivating example. To infer the current time, we must first identify the clue words `10:45' in $u_6$ and `15 minutes' in $u_7$. Then we must conduct a logical inference based on these clue words in $u_6$ and $u_7$.

To tackle these challenges, first, we need better contextual representation for identifying the clue words in conversations. This is because clue word identification inevitably relies on the context of a conversation. Although previous literature publications have achieved promising results in context modeling~\cite{Tao2019,qu2019bert,su2019improving}, there are still several limitations of these approaches. More concretely, the existing studies either concatenate the utterances to form context~\cite{su2019improving} or process each utterance independently~\cite{Tao2020,Lu2020}, leading to the loss of dependency relationships among utterances or important contextual information. It has been validated that the chronological dependency between utterances~\cite{Lu2020,yeh2019flowdelta}, as well as the semantical dependency between utterances, are crucial for multi-turn response selection. Thus, how to model the dependencies in utterances remains a challenging problem for context representation.

Second, we need to devise a new strategy to collect the clue words scattered in multiple utterances and need to reason according to these clue words. In recent years, we have witnessed great success in KBQA (knowledge base question answering) and MRC (machine reading comprehension) tasks. However, new obstacles emerge for transferring current reasoning approaches in KBQA and MRC to conversational reasoning. (i) A clear reasoning path based on entities in a well-structured knowledge base exists in KBQA~\cite{2017Variational,xu2019enhancing}, but there is no similar reasoning path in utterances. (ii) Current approaches on MRC conduct inference based on graph while taking shared entities as nodes~\cite{Fang2019,Qiu2020,Martin}, while it is difficult to construct such graphs based on entities in short utterances, which usually suffer from greater coreference resolution, poor content and serious semantic omission problems in comparison with document text.

In this paper, we propose a new model named GRN (graph reasoning Network) which can tackle both challenges in an end-to-end way. We first introduce two pre-training tasks called NUP (next utterance prediction) and UOP (utterance order prediction) which are specially designed for response selection. NUP endows GRN with context-aware ability for semantical dependency, and UOP facilitates GRN with the ability to capture the chronological dependency. These customized pre-training methods are beneficial for modeling dependencies contained in utterances to achieve better context representation. We perform task-adaptive pre-training with the combined NUP and UOP tasks based on the ALBERT model~\cite{Lan2020}. To conduct reasoning based on clue words, we devise a graph neural network called UDG (utterance dependency graph), which not only models the dependencies between utterances with each utterance as a node but also collects the clue words from different utterances. Reasoning is achieved by propagating the messages of clue words between nodes along various utterance paths on UDG, and this graph reasoning structure realizes the inference based on an utterance-level context vector with local perspective. On the other hand, we also implement a reasoning network by the output of the trained model and self-attention mechanism. This sequence reasoning structure realizes the inference based on the highly summarized context vector with global perspective.
To summarize, we make the following contributions:

\begin{itemize}
    \item We introduce two customized pre-training methods NUP and UOP, which are specially designed for response selection.
    \item We propose a graph neural network, UDG, which can capture the clue word dependency between utterances and realize reasoning by propagating messages along various utterance paths.
    \item We integrate a graph reasoning network with a self-attention based sequence reasoning network in an end-to-end framework. The empirical results show that our proposed model outperforms the strong baseline methods by large margins on both MuTual and $\text{MuTual}^{plus}$ datasets.
\end{itemize}

\section{Related Work}
\subsubsection{Response Selection}
Response selection aims to select the best matched response from a set of candidates, which can be categorized into single-turn and multi-turn dialogues. Early studies focused on the single-turn dialogues. Recently, researchers devote more attention to the multi-turn dialogues technology ~\cite{Tao2019,Tao2020,Lu2020}. Existing methods tend to use deep matching methods to model the relationships between utterances and candidate responses. These models generally use representation methods based on LSTM, attention mechanisms and hierarchical interaction techniques. These models are focused on \textbf{matching} not \textbf{reasoning}. The key problem of matching type models is how to extract better matching features.  In fact, however, the key problem of reasoning is how to conduct inference according to clue words from different utterances, which is more complicated. Existing multi-turn response selection methods are not suitable for the reasoning problems.
\subsubsection{Graph Neural Network}
The GNN (graph neural network) has achieved outstanding performance in reasoning tasks based on question answer ~\cite{Qiu2020}. In this paper, we convert the sequence structure of utterances into a graph structure and realize reasoning by using the GCN (graph convolutional network). The graph structure network is adept at information collection, fusion and summarization by message passing along different node paths. The reasoning problem of dialogue is different from those of KBQA and MRC. There is a clear reasoning path based on entity in triples knowledge in KBQA. Currently, most applications of MRC construct the graph according to the shared entities. All of these approaches are difficult to perform for conversation. Previous works on GCN show the superior ability of GCN to integrate the features of local nodes, which inspires us to solve the inference issues in conversation with graph structure.
\section{Methodology}
\subsection{Problem Formalization}
Given a conversation context $U=\{u_1,u_2,...,u_n\}$ where $u_i=\{w_1,w_2,...,w_{l_i}\}$ with $l_i$ tokens represents an utterance, $R=\{r_a,r_b,r_c,r_d\}$ and $y \in \{a,b,c,d\}$ is a four-category label, indicating which $r_{i \in \{a,b,c,d\}}= \{w_1,w_2,...,w_{|r_i|}\}$ is a proper response for $U$. Our goal is to learn a matching model $f(U,r_i)$, which can measure the relevance between the context $U$ and each candidate response $r_{i \in \{a,b,c,d\}}$.
\begin{figure*}[t]
\centering 
\includegraphics[width=0.98\textwidth]{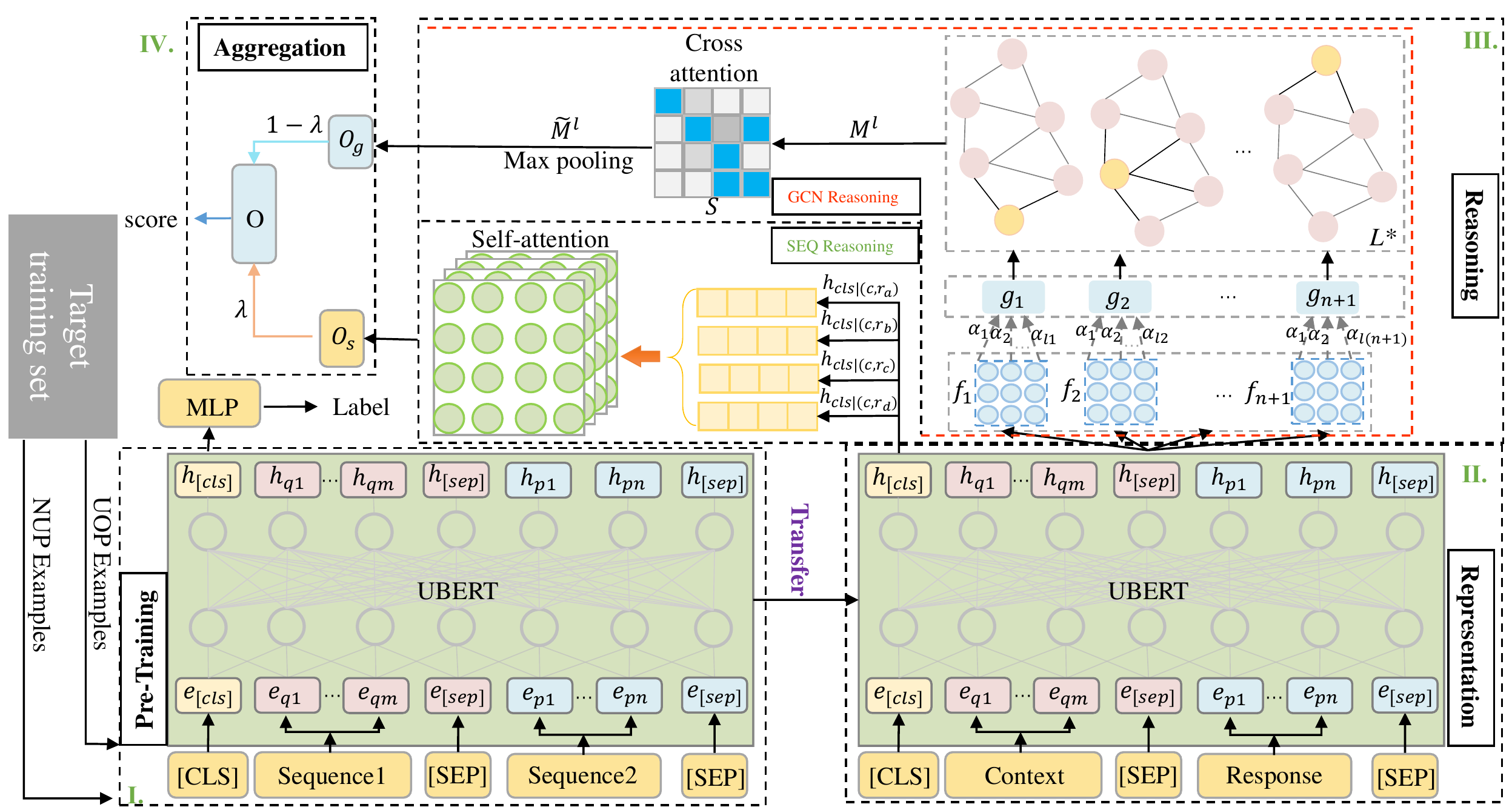}
\caption{Overview of our GRN. The UBERT on the left is pre-trained with the combining tasks using NUP and NOP based on the ALBERT. We fine-tune UBERT (right) on the downstream task.}
\label{framework}
\end{figure*}
\subsection{Model Overview}
Figure~\ref{framework} shows an overview of GRN, which follows the \textbf{Pretraining-Representation-Reasoning-Aggregation} framework. First of all, we perform unsupervised pre-training on the target training set with combined NUP and UOP tasks on ALBERT. 
We refer to the pre-trained model as UBERT. We then fine-tune UBERT on downstream tasks. For each input pair $(U,r_i)$, the output of UBERT is denoted as $H \in R^{L*d}$, where $L=\sum_{i=1}^n l_i + |r_j|+3$ and $d$ is the dimension size of the output layer UBERT. We obtain the summary vector $h_{[cls]}$ for input pair $(U,r_i)$ and token-level representations for all the utterances and the candidate responses $r_i$ from $H$ are denoted as $f=\{f_1,f_2,...,f_{n+1}\} \in R^{(n+1)*d}$.

The reasoning module then starts to compute the semantic relevance between context and response at different granularities through sequence and graph network structure. The sequence reasoning module calculates the semantic relevance based on summary information from the global perspective, and the graph reasoning module computes the semantic relevance based on contextual information from local perspective. Features from different modules
are merged through the aggregation layer. Next, we will introduce in detail the implementation of every module.

\subsection{Pre-training}
Continued pre-training on the target datasets can improve the performance of the model ~\cite{Gururangan2020}. Motivated by this, we follow the previous works on the target training set based on ALBERT. Unfortunately, the performance of the model has suffered significant degradation, for thich there are two main reasons.

\begin{itemize}
    \item As in the previous analysis ~\cite{Gururangan2020}, the domain of our target datasets is included in the original ALBERT pre-training domain.
    \item The form of original pre-training tasks is different from that of our target task.
    \item We demand a customized pre-trained language model for multi-turn response selection.
\end{itemize}

To solve these problems, we devise two auxiliary pre-training tasks, UOP and NUP. They capture dependencies between utterances from different perspectives. Next we introduce the two tasks in detail.

\subsubsection{Utterance order prediction (UOP)} Coherence and cohesion in discourse have been widely studied ~\cite{grosz1995centering}. Existing works mainly learn sentence representation from token-level and sentence-level. The token-level methods~\cite{kiros2015skip,hill2016learning} learn sentence embedding by predicting tokens in neighboring sentences. Methods based on sentence-level ~\cite{gan2017learning,Lan2020} learn sentence representation by prediction of the future sentences. Unlike most of the above work, we consider all utterances as a whole. We want our model to identify which utterance sequences are ordered and which are disordered. Concretely, the ordered sequences are labeled as pos-seq and the randomly shuffled utterance sequences are labeled as neg-seq.

\subsubsection{Next utterance prediction (NUP)} It is a fact that response selection is heavily dependent on the previous dialogue segments. Responses differ when giving different dialogue segments. In other words, there is a semantic dependency between previous dialogue segments and the next utterance. To capture the semantic dependency, we demand our model to distinguish the previous dialogue segments from the future dialogue segments for an utterance. Concretely, in choosing an utterance $u_i$ at random from utterances $U$, where $1 < i < n $, the left part of $U$ is denoted as $U_{left}=\{u_1,u_2,...,u_{i-1}\}$ and right part is denotes as $U_{right}=\{u_{i+1},u_{i+2},...,u_n\}$. We require the model to distinguish the $(U_{left},u_i)$ pair from the $(U_{right},u_i)$ pair. 

\subsection{Representation}
This module uses the pre-training UBERT to encode the utterances $U$ and $R$ generating the contextual representations $H$. First, we concatenate the utterances $U$ into a sequence denoted as $T$. For each sequence pair $(T,r_i)$, we generate an input to feed through UBERT by concatenating $[CLS] + T + [SEP] + r_i + [SEP]$. At the same time, we employ $E_i=\{s_1,e_1,s_2,e_2,...,s_{n+1},e_{n+1} \in R^{2n+2}\}$ to store the start and end positions of each utterance including the response candidate $r_i$. The output $h_{[cls]}$ from UBERT is considered as a summary vector for each sequence pair $(T,r_i)$.

\subsection{Reasoning}
The goal of the reasoning layer is to calculate semantic relevance between utterances and each candidate response. We design two different structure networks to compute the relevance from different perspectives. The sequence structure network reasoning module is responsible for calculating semantic relevance from the global perspective. The graph structure network reasoning module is responsible for calculating local semantic relevance.

\subsubsection{Sequence structure Reasoning}
The representation $h_{[cls]}$ of the special token $[CLS]$ is often used for classification, which is considered as the summary vector for the input sequence pair. One common way is to project this summary vector into a scalar ~\cite{devlin2018bert,Qiu2020}, which computes a score independently for each response. In this paper, we consider selecting the most suitable response by comparing four candidates responses. Therefore, we add a multi-head self-attention ~\cite{Vaswani2017} layer on top of the concatenation of all $[CLS]$. The self-attention is defined as:

\begin{equation}
Multihead=Concat(head_1,...,head_h) W^O
\end{equation}
\begin{equation}
head_i=Attention(Q W_i^Q,KW_i^K,VW_i^V)
\end{equation}
\begin{equation}
Attention(Q,K,V)=softmax(\frac{QK^T}{\sqrt{d}}) V
\end{equation}

where $W_i^Q$, $W_i^K$ and $W_i^V$ are linear projection matrices, and $h$ is the number of the heads. The motivation for adding a multi-head self-attention accross the special token $[CLS]$ generated from different candidate responses with utterances is to encourage the further interaction between utterances and candidate responses by comparing four candidate responses, which can generate a better representation for selecting a proper response. The output of this layer is defined as $O_s$.

\subsubsection{Graph Structure Reasoning}
The GCN (Graph Convolutional Network) has been proven to offer better performance for the reasoning task in QA ~\cite{Martin,Fang2019,Qiu2020}. The graph structure allows messages to pass over the nodes with local contextual information, which can comprehensively consider clue words in different utterances to achieve the purpose of reasoning. Previous works in QA show that GCNs have achieved promising results in reasoning tasks. The reason GCN works is that GCN can summarize the feature information of local nodes. In this paper, we employ GCN to summarize the local feature information. We propose to build a graph model over utterance representations $g_i$ to explicitly facilitate reasoning over all utterances. The graph node is initialized by the utterance representations $g_i$.

To build the connections among graph nodes, we design two types of edges based on the dependencies between utterances. We named the constructed graph UDG, taking utterances as nodes.

\begin{itemize}
    \item Add an edge between adjacent utterances.
    \item Add an edge between two nonadjacent utterances if there is a dependency between topics to which they belong. The topic is an abstract concept. Each utterance has a central idea which is called a topic in this paper. For convenience of description, we use a highly abstract token or phrases to represent the topic. However, in our implementation, the topic is composed of a keyword or a set of highly correlated keywords for the utterance. 
    It is a fact that the topic of the utterance will change frequently over time. However, changes in topics are always logically related. That is, there is logical dependence between topics. In this paper, we extract the named entities and keywords by TextRank~\cite{mihalcea_textrank}. We then use community detection algorithms  to determine whether there is dependency between topics.
\end{itemize}

The motivation for the first type of edge is that we want the GNN to grasp the chronological dependency among utterances. Furthermore, we design the second type of edges according to the logical dependency among topics. Cross-utterance reasoning is realized by jumping from one utterance to another one with different topics.

\begin{figure}[t]
\centering
\includegraphics[width=0.98\columnwidth]{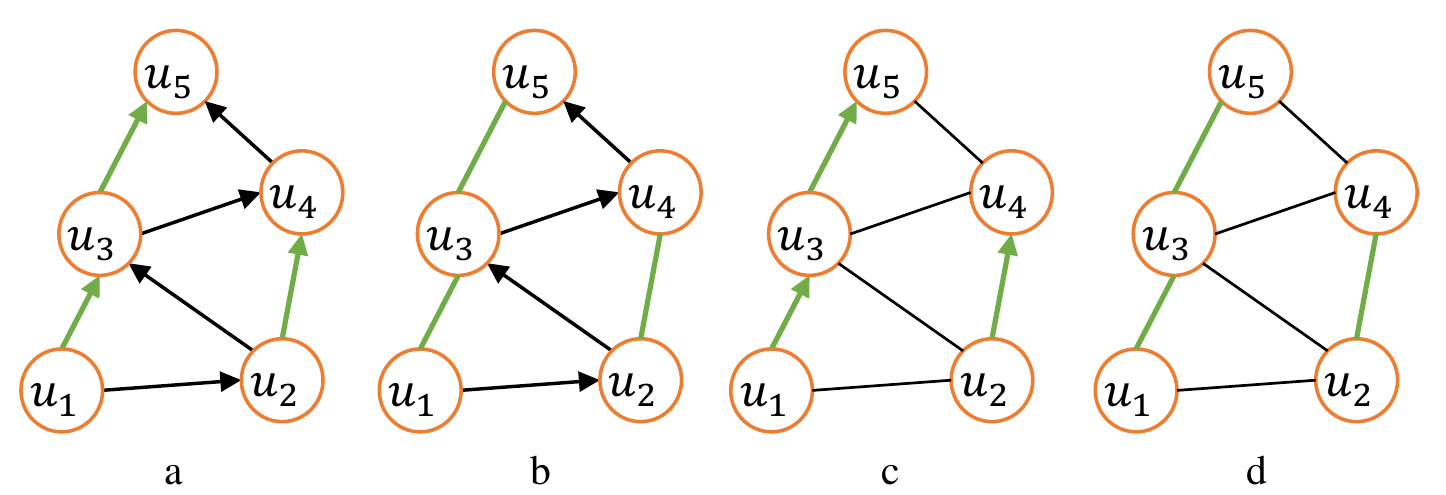}
\caption{Different types of UDG. The black line represents the chronological dependency relationship between utterances. The green line represents the dependency between topics.}
\label{dag}
\end{figure}

It is worth noting that the last utterance is not always the most important, and previous utterances may contain key information. Therefore, it is not always reasonable to denote the two edges as directed. When using a directed edge, we should pledge that information of all nodes can reach the last node along the specified direction, which corresponds to the long distance dependency issue. An undirected edge allows message to pass in both directions between two nodes. Therefore, we employ undirected UDG as the basic graph reasoning structure. To deeply study the impact of different UDGs on performance, we will discuss the rationality of all possible UDGs in detail and present the corresponding experimental results. All possible UDGs of the combinations of edges in different directions are shown in Figure~\ref{dag}.

Given the output $H$ from the UBERT on the downstream task, we first obtain sequential representation of each utterance from $H$.
\begin{equation}
    f_i=H[s_i:e_i,:] \in R^{l_i * d}
\end{equation}
where $s_i$ and $e_i$ represent the start and end positions of utterance $u_i$ in $H$, respectively. For each utterance, we adopt the attention mechanism to dynamically select clue words, which is defined as:
\begin{equation}
    score_j=v_1^T tanh(f_i^j W_1 + f_i^k W_2)
\end{equation}
\begin{equation}
    \alpha_j=\frac{exp(score_j)}{\sum_k exp(score_k)}
\end{equation}
\begin{equation}
    g_i=\sum_{j=1}^{l_i} \alpha_j f_i^j
\end{equation}
where $g_i \in R^d$, $W_1,W_2,v_1$ are trainable parameters.

Now let us briefly describe the GCN ~\cite{kipf2016semi} layers used in Figure~\ref{framework}. In general, the input of the GCN is a graph $G=\{V,E\}$ with $n$ vertices $v_i \in V$, edge $e_{ij}=(v_i,v_j) \in E$ with $w_{ij} = 1$ , and $w_{ij} = 0$ if $e_{ij}=(v_i,v_j) \not\in E$. We use the adjacency matrix $A \in R^{n*n}$ of the graph to represent the weight of edges. In addition, the input of the graph contains a vertex feature matrix denoted as $ M^0 \in R^{n*d}$. In this paper, suppose \textit{D} is a diagonal matrix such that $D_{ii}=\sum_j A_{ij}$, $M_i^0$ represents initial graph node embedding from utterance representation $g_i$, and $M^l \in R^{n*d_l}$ denotes the hidden representations of the vertices in the $l^{th}$ layer, where the calculation of node representations after each layer can be formulated as:
\begin{equation}
  M^{l}=\sigma(\tilde{D}^{-\frac{1}{2}} \tilde{A}\tilde{D}^{-\frac{1}{2}} M^{l-1} W^l)
\end{equation}
where $\tilde{A}=A+I_N$, $I_N$ is the identity matrix, and $\tilde{D}$ is a diagonal matrix with $\tilde{D}_{ii}=\sum_j \tilde{A}_{ij}$ . These are the adjacency matrix and the degree matrix of graph \textit{G}, respectively. $W^l$ represents the trainable parameters in the $l^{th}$ layer. $\sigma(.)$ denotes the ReLU (rectified linear unit) activation function. After message passing on the UDG, each node has a local contextual representation. However, indirect weak interaction is determined for nonadjacent nodes by a graph convolutional rule. Therefore, we use the cross attention mechanism to discover a further interact between utterances, which encourages the model to reason based on global information. The calculation formula is defined as follows:
\begin{equation}
  S_i^j=v^T tanh(M_i^l W_l + M_j^l W_r + M_i \odot M_j)
\end{equation}
\begin{equation}
  \alpha_i^j=\frac{exp(score_i^j)}{\sum_j score_i^j}
\end{equation}
\begin{equation}
  \tilde{M}_i^l=\sum_j \alpha_i^j * M_j^l
\end{equation}
where $W_l,W_r,v^T$ are trainable parameters, and $S$ is the similarity matrix across utterances. Finally, the hidden representations of all nodes are merged into a single vector of a fixed length, denoted as $O_g$, by utilizing the max-pooling operation of the hidden vectors of all vertices.
\subsection{Aggregation}
An aggregation layer aggregates the sequence matching result $O_s$ and the graph matching result $O_g$ by the gate attention mechanism, which is defined as:
\begin{equation}
    g=sigmoid(W_g [O_s;O_g])
\end{equation}
\begin{equation}
    O=g*O_s + (1-g) * O_g
\end{equation}
where $W_g$ is a trainable parameter. We finally compute matching score based on the mixed matching features $O$ via a single layer perceptron, which is defined as:
\begin{equation}
    score_{r_i}=W_o * O + b_o \quad i \in \{a,b,c,d\}
\end{equation}
where $W_o, b_o$ are trainable parameters. The loss function is negative log likelihood, defined as:
\begin{equation}
    p_i=\frac{exp(score_{r_i})}{\sum_j exp(score_{r_j})}
\end{equation}
\begin{equation}
    L=-\sum_i y_i^{oh} log(p_i)
\end{equation}
where $y^{oh}$ is the one-hot vector of the real label $y$.

\section{Experiments}
\begin{table}[t]
\centering
\begin{tabular}{p{1.9cm}|p{0.6cm}|p{0.6cm}|p{0.6cm}|p{0.6cm}|p{0.6cm}|p{0.6cm}}  %
\hline
   & \multicolumn{3}{c|}{MuTual} & \multicolumn{3}{c}{$ \text{MuTual}^{plus}$} \\ \hline
   Method & R@1 &R@2 & MRR & R@1 &R@2 & MRR \\ \hline
   Human & 93.8 & 97.1 & 96.4  & 93.0 & 97.2 & 96.1 \\ \hline
   Random  & 25.0 & 50.0 & 60.4 & 25.0 & 50.0 & 60.4  \\ \hline
   TF-IDF & 27.9 & 53.6 & 54.2 & 27.8 & 52.9 & 76.4  \\
   DuLSTM & 26.0 & 49.1 & 74.3 & 25.1 & 47.9 & 51.5  \\
   SMN & 29.9 & 58.5 & 59.5 & 26.5 & 51.6 & 62.7 \\
   DAM & 24.1 & 46.5 & 51.8 & 27.2 & 52.3 & 69.5  \\ \hline
   BIDAF & 35.7 & 58.9 &58.9 & 33.4 & 49.2 & 56.2 \\
   RNET & 27.0 & 43.5 & 51.3 & 26.1 & 50.6 & 53.2 \\
   QANET &24.7 &51.7  & 52.2 & 25.1 & 49.5 & 51.9 \\ \hline
   BERT  & 64.8 & 84.7 & 79.5 &  51.4 & 78.7 & 71.5 \\
   RoBERTa & 82.5 & 95.3 & 90.4 & 75.7 & 92.8 & 85.6\\
   SpanBERT & 80.6 & 94.8 & 89.3 & 70.3 & 88.4 & 83.0\\
   GPT-2 & 33.2 & 60.2 & 58.4  & 31.6 & 57.4 & 56.8  \\
   GPT-2-FT  & 39.2 & 67.0 & 62.9 & 22.6 & 61.1 & 53.5  \\
   BERTMC  & 66.7 & 87.8 & 81.0 & 58.0 & 79.2 & 74.9 \\
   RoBERTaMC & 68.6 & 88.7 & 82.2 & 64.3 & 84.5 & 79.2  \\
   ALBERT & 84.7 & 96.2 & 91.6 &  78.9 & 94.6 & 88.4 \\ \hline
   GRN  & \textbf{91.5} & \textbf{98.3} & \textbf{95.4} & \textbf{84.1} & \textbf{95.7} & \textbf{91.3} \\ \hline
\end{tabular}
\caption{Experimental results of different methods on two testing sets}
\label{compared_results}
\end{table}

\subsection{Datasets}
We test our proposed GRN on MuTual and $\text{MuTual}^{plus}$. MuTual is built based on Chinese high school English listening comprehension test data, consisting of 8,860 challenge questions, in terms of almost all questions involving reasoning, which are designed by linguistic experts and professional annotators. MuTual consists of 8,860 context-response pairs and has an average of 4.73 turns. Each context-response pair has four candidate responses. $\text{MuTual}^{plus}$ is built based on MuTual by using \textbf{safe response}  to replace one of the candidate responses for each instance in MuTual. $\text{MuTual}^{plus}$ is used to check whether the model can select a safe response when the other candidates are incorrect.

\subsection{Metrics}
We use the same evaluation metrics as those used in previous works ~\cite{mutual2020}. Each compared model must give the recall at position 1 in 4 candidates(R@1), recall at position 2 in 4 candidates(R@2) and Mean Reciprocal Rank (MRR) ~\cite{Ricardo2016}.
\subsection{Baselines}
\begin{table}[t]
\centering
\begin{tabular}{l|l|l|l}
\hline
   Method & R@1 & R@2 & MRR \\  \hline
    GRN & 93.5 & 98.5 & 97.1 \\
   -pre-training  & 90.5 & 97.3 & 94.7 \\
   -GCN match & 91.5 & 97.9 & 95.5 \\
   -sequence match & 91.3 & 97.6 & 95.2  \\
   -cross attention & 92.2 & 97.3 & 95.6 \\
   -SelfAtt ~\cite{Vaswani2017} & 92.7 & 98.2 & 96.8  \\ \hline

\end{tabular}
\caption{Ablation experimental results of GRN on MuTual validation set}
\label{ablation_result}
\end{table}
\textbf{Basic Models:} Models in ~\cite{2015Lowe} include TF-IDF and DuLSTM. \textbf{SMN} ~\cite{Wu2017}: A matching model calculates the relevance feature on token-level. \textbf{DMN} ~\cite{Zhou2018}: A matching model
calculates the relevance based on semantic and functional dependencies by using stacked self-attention ~\cite{Vaswani2017}. \textbf{MRC Methods}: We test the performances of several representative MRC models, QANET ~\cite{yu2018qanet}, BIDAF ~\cite{seo2017bidirectional} and R-NET ~\cite{wang2017gated}. \textbf{BERT} ~\cite{devlin2018bert}: An autoencoding language model based on transformer.  \textbf{SpanBERT} ~\cite{joshi2020spanbert}: An autoencoding language model with span masking base on transformer. \textbf{RoBERTa} ~\cite{2019RoBERTa}: An autoencoding language model with dynamic masking base on transformer. \textbf{ALBERT} ~\cite{Lan2020}: An improved language model base on BERT. \textbf{GPT-2} ~\cite{Radford2018}: We fine-tune the GPT-2 on (context,response) pair and choose the one with the lowest perplexity as the correct response. In addition, we fine-tune the GPT-2 only using the positive response and context to construct the input sequence, whcih is denoted as \textbf{GPT-2-FT}. \textbf{Multi-choice Method}: Different from the previous work ~\cite{devlin2018bert}, we concatenate the CLS representations of all input sequences in one instance to calculate matching score, which is denoted as BERTMC. This method is also applicable to other language models similar to BERT, such as RoBERTa.
\begin{table}[t]
\centering
\begin{tabular}{l|l|l|l}
\hline
   Pre-training Method & R@1 & R@2 & MRR \\  \hline
   ALBERT(original) & 84.8 & 96.0 & 91.6  \\
   BERT ~\cite{devlin2018bert} & 78.2 & 90.6 & 84.9  \\
   Our Pre-training  & \textbf{87.6} & \textbf{96.7} & \textbf{93.6}  \\  \hline

\end{tabular}
\caption{Performance comparison of UBERT using different pre-training methods on the validation dataset}
\label{pre-training_result}
\end{table}

\subsection{Implementation Details}
\subsubsection{Unsupervised Pre-training} In this paper, we use ALBERT ~\cite{Lan2020} as the base model. We construct the train corpus based on the MuTual train dataset without using the response candidate. We set
the initial learning rate as 3e-5, the train step is 120,000. The trained model is called UBERT, which was trained based on the UOP and NUP task. The accuracy of unsupervised pre-training achieves 0.98.
\subsubsection{Fine-tune}
In the downstream task, we use both the sequence and the graph structure network and use pre-trained U-BERT to initialize downstream model parameters. We set the learning rate as 2e-5, the number of GCN layers as $l=2$ and use Adam optimizer to update the model parameters. The hidden size of GRN is 512. We train the model for 3 epochs, and the best performance on the validation set was considered as the final model.

\begin{table}[t]
\centering
\begin{tabular}{l|l|l|l}
\hline
   UDG Type(Figure\ref{dag}) & R@1 & R@2 & MRR \\  \hline
   Type a  & 88.6 & 93.2 & 91.9  \\
   Type b  & 90.3 & 96.7 & 94.2  \\
   Type c  & 91.8 & 97.9 & 95.5  \\
   Type d & \textbf{93.5} & \textbf{98.5} & \textbf{97.1}  \\ \hline

\end{tabular}
\caption{Performance comparison of different UDGs on the validation dataset}
\label{dag_result}
\end{table}

\begin{table}[t]
\centering
\begin{tabular}{l|l|l|l|l|l}
     \hline
      Model & T=2 & T=3 & T=4 & T=5 & T $\geq$ 6  \\ \hline
     Instances & 290 & 143 & 115 & 51 & 287  \\
     RoBERTa & 73.1 & 65.7 & 63.5 & 80.4 & 71.2  \\
     RoBERTa-MC & 68.1 & 62.2 & 60.9 & 72.5 & 75.0  \\
     ALBERT & 85.6 & 82.1 & 82.5 & 84.3 & 86.0  \\
     GRN & \textbf{92.1} & \textbf{93.1} & \textbf{88.6} & \textbf{88.2} & \textbf{91.9} \\  \hline
\end{tabular}
\caption{R@1 performance comparison of different number of turns on the test set. T denotes number of
turns.}
\label{acc_results}
\end{table}
\subsection{Results}
Table~\ref{compared_results} reports the testing results of GRN as well as all comparative models on MuTual and $\text{MuTual}^{plus}$. We can observe that the performance of GRN significantly outperforms all comparative models on both datasets, demonstrating the superior power of GCN in reasoning questions with multi-turn context. One notable point is that the performance of traditional representation models (i.e., TF-IDF, DuLSTM, SMN and DMN) is relatively low. This indicates that these representation models have insufficient reasoning ability. Compared with ALBERT, GRN has an absolute advantage of $6.8\%$ on R@1, approximately $2\%$ on R@2 and approximately $3.8\%$ on MRR on MuTual. Moreover, nearly the same performance improvement is observed on $\text{MuTual}^{plus}$, which again verifies that our reasoning strategy is effective. The performances of language models such as BERT, RoBERTa, etc. cannot compete with our pre-training strategy on both datasets, which demonstrates that the original language model cannot capture the rich context representations better in conversation.
We also report the testing results of some classic MRC models such as QANET, BIDAF and R-NET. Their performances are very close to those of the previous representation models. The main reason for this is that these models have limited reasoning capabilities.

\subsection{Ablation Study}
We investigate the effects of different parts of the GRN through removing them one by one from GRN, as shown in Table~\ref{ablation_result}. \textbf{-pre-training}: removing our pre-training strategy. The performance of the -pre-training method drops greatly, which verifies the effectiveness of pre-training in capturing dependencies among utterances. \textbf{-GCN reasoning}: removing the GCN reasoning module including the graph attention. the -GCN reasoning method causes considerable performance degradation, demonstrating the superior power of GCN in reasoning tasks. \textbf{-sequence reasoning}: removing the sequence reasoning module including the multi-head attention. The -sequence reasoning method also causes notable performance degradation. The sequence reasoning module can capture the information of clue words from the global perspective by deep interaction between different (utterance,response) pairs. \textbf{-cross attention}: removing the attention on the top of GCN and using max-pooling for the hidden vectors of all vertices. The performance of the -graph biattention method drops slightly, which demonstrates that the attention mechanism can encourage the interaction between nonadjacent nodes on UDG. \textbf{-SelfAtt}: removing the multi-head self-attention on top of the CLS token. There is also a slight decrease in performance, which demonstrates that the self-attention module is beneficial for capturing the key information.

\subsubsection{Pre-training method discussion}
Table~\ref{pre-training_result} reports the results of GRN with different pre-training methods. As demonstrated, training ALBERT on the target training set following the previous work ~\cite{Lan2020} results in a significant drop in performance, which verifies our previous analysis that the domain of our dataset is similar to the original ALBERT pre-training domain. The performance of UBERT exhibits a significantly improvement, whcih demonstrates the effectiveness of our pre-training strategy.
\begin{figure}[t]
\centering
\includegraphics[width=1\columnwidth]{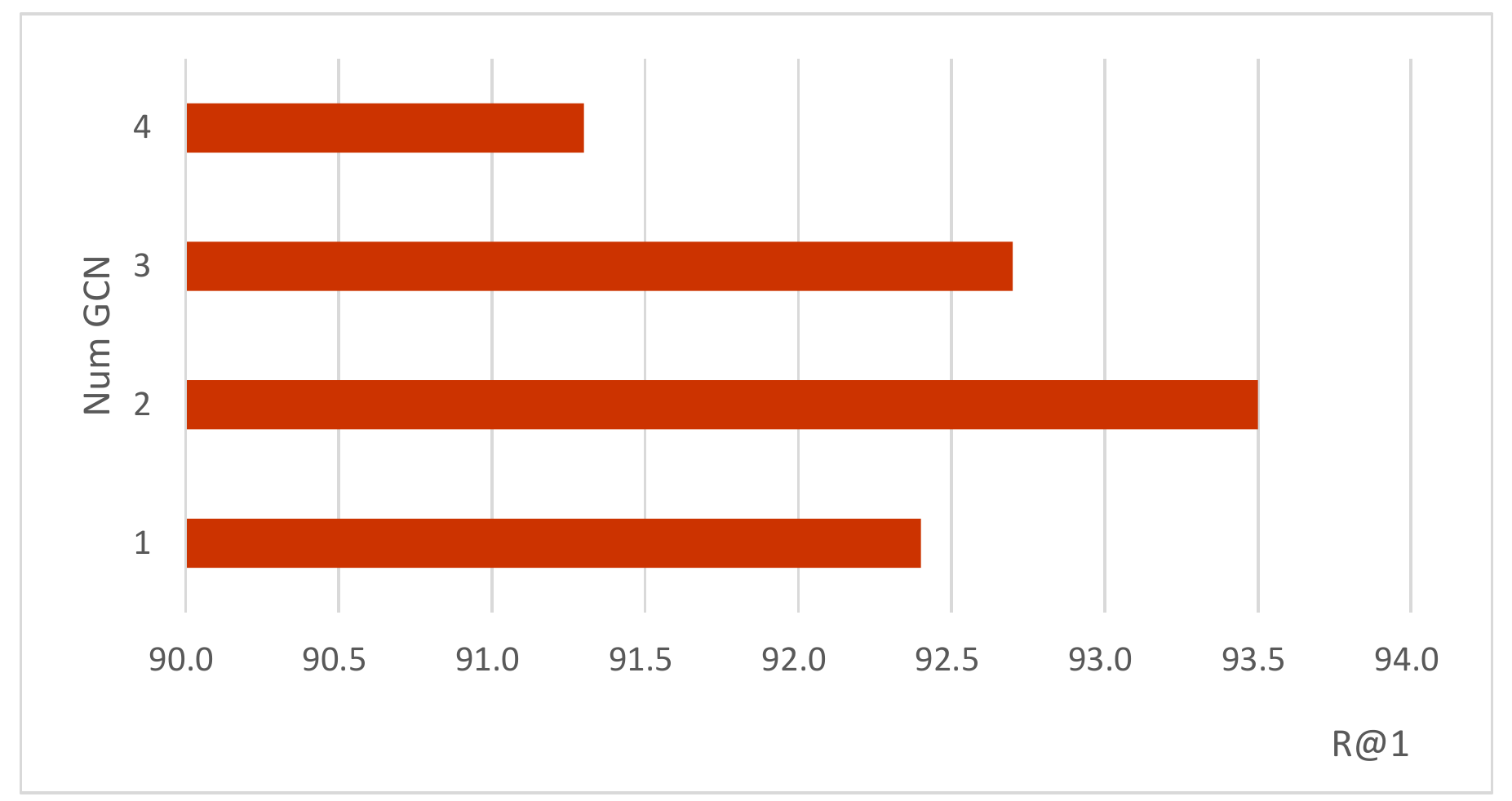}
\caption{R@1 performance comparison of different number of GCN layers on validation dataset.}
\label{NGCN}
\end{figure}
\begin{figure*}[t]
\centering 
\includegraphics[width=0.98\textwidth]{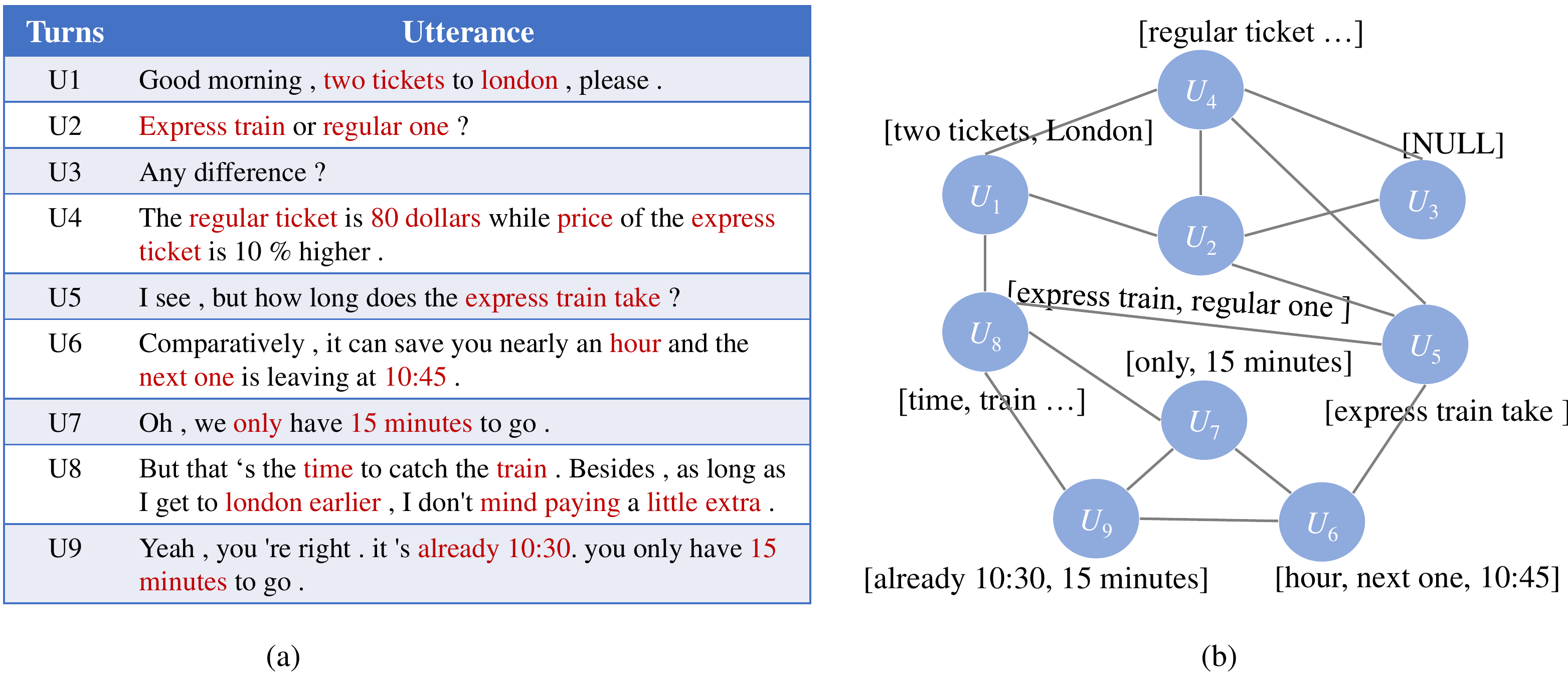}
\caption{Case Study. (a) is one of the keyword examples for utternces in Figure~\ref{dataem} and red tokens is the keywords in utterance. (b) is the corresponding UDG, the tokens is the keywords for every utterance.}
\label{case}
\end{figure*}
\subsubsection{Number of GCN layers discussion}
 Figure~\ref{NGCN} shows the effect of GCN layers on model performance. It can be found that the accuracy(R@1) of the GRN is highest when the number of GCN layers is 2. An interesting phenomenon is noted in that deeper numbers of GCN layers do not correspond with better performance of the model. As analyzed in previous work~\cite{klicpera2018predict},the number of GCN layers is related to the depth of the graph and the sparseness of the adjacency matrix. The number of turns in Mutual is mostly within 5, which makes it more appropriate for GRN to use shallow GCN.
\subsubsection{UDG discussion}
To investigate the effect of different UDGs on performance, we test the performance of GRN using different UDGs on the validation dataset. Table~\ref{dag_result} reports the resuts of the model using different UDGs. As demonstrated, the performance of GRN based on undirected UDG significantly outperforms the others. The GRN based on directed UDG yields the lowest performance. We can conclude that it is not an effective method for ensuring that the message passes along the chronological order of utterances, which requires the pledge that all key information can reach the last node by message passing. In fact, according to instance analysis, each utterance may contain clue words and reasoning which is not developmental in one direction. Therefore, reasoning is more realistic for an undirected UDG.
\subsubsection{Performance across different context lengths}
We investigate the effects of different turns on the performance of GRN. Table~\ref{acc_results} demonstrates the performance of GRN on MuTual. As demonstrated, the performance of GRN outperforms all the compared methods on different turns of MuTual. It is notable that the performance of GRN does not decrease significantly as the number of turns increase, demonstrating the good adaptability of the GRN to different turns.
\subsection{Case study}
Figure~\ref{case} (left) shows a case of keyword extraction in a dialogue and the right part is the UDG corresponding to this example. As described before, the topic of each conversation is an abstract concept. In this paper, the topic is represented by keywords of utterance. As can be seen from the Figure~\ref{case} (left), the keywords set approximate represents the main meaning of each sentence. Then, based on the extracted keywords, we use a community detection algorithm to discover the relationship between non-adjacent utterances. When there is not keyword included in utterance, we use the special token NULL instead.
\begin{figure}[t]
\centering
\includegraphics[width=1\columnwidth]{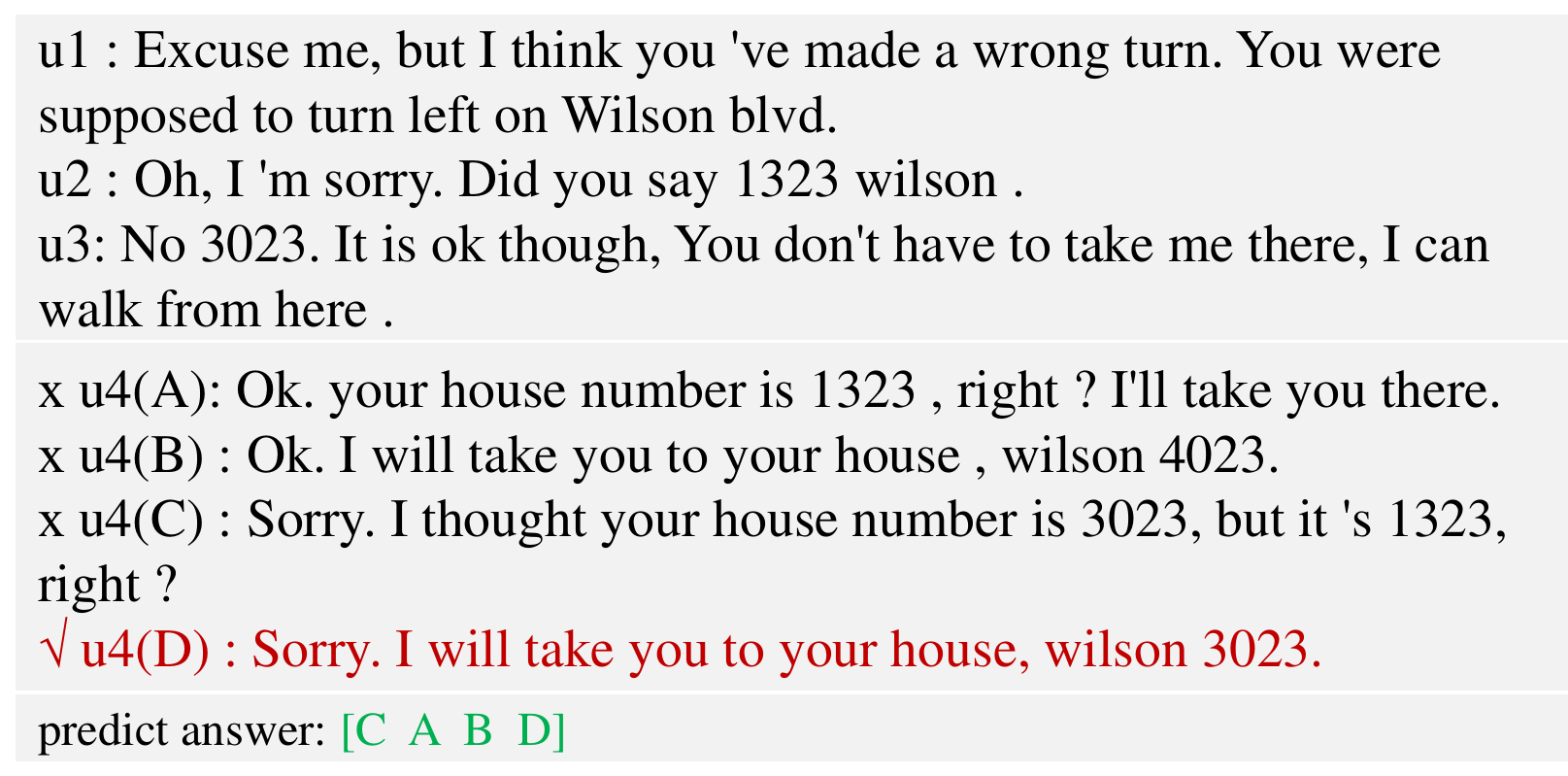}
\caption{Error analysis}
\label{error}
\end{figure}
\subsection{Error analysis}
Although GRN outperforms all baseline methods on the two datasets, there are still some problems which cannot be dealt with.

\begin{itemize}
    \item Logical Reasoning. Although the GRN can handle some basic reasoning problems, it cannot handle more complex logical reasoning with complex grammatical phenomena. As demonstrated in Figure~\ref{error}, the GRN cannot understand the semantic and logical relationships between `NO 3023' and previous utterances. The GRN thinks `1323' is the correct number and "3023" is wrong according to the predicted answer order. The utterance `NO 3023' gives its own view while negating the previous conclusion. In addition, this utterance omits substantial information, where is the dialogue is more difficult to handle than the general document.
    \item Safe Response. The performance on $\text{MuTual}^{plus}$ is significantly lower than that on MuTual. This demonstrates that the GRN cannot select the safe response when other candidate responses are incorrect in some cases.
\end{itemize}

\section{Conclusion}
In this paper, we propose a new architecture for multi-turn response reasoning. Concretely, we first propose NUP and UOP pre-training tasks for response selection. We design the UDG of utterance for reasoning. We introduce sequence and graph reasoning structure jointly, where the sequence reasoning module can capture the key information from the global perspective and the graph reasoning module is responsible for capturing the clue words information from the local perspective. The experiment results on MuTual and $\text{MuTual}^{plus}$ achieve a new heights. There is still expansive room for improvement in performance on $\text{MuTual}^{plus}$. In future work, we will further investigate how to balance safe response and meaningful candidate response.
\section{Acknowledgement}
The project is supported by the National Key R\&D Program of China (2018YFB1004700), the National Natural Science Foundation of China (61872074, 61772122, U1811261), and the Fundamental Research Funds for the Central Universities (No.N181602013).
\bibliography{reference}

\begin{thebibliography}{33}
\providecommand{\natexlab}[1]{#1}
\providecommand{\url}[1]{\texttt{#1}}
\providecommand{\urlprefix}{URL }
\expandafter\ifx\csname urlstyle\endcsname\relax
  \providecommand{\doi}[1]{doi:\discretionary{}{}{}#1}\else
  \providecommand{\doi}{doi:\discretionary{}{}{}\begingroup
  \urlstyle{rm}\Url}\fi

\bibitem[{{Baeza-Yates} and {Ribeiro-Neto}(1999)}]{Ricardo2016}
{Baeza-Yates}, R.~A.; and {Ribeiro-Neto}, B. 1999.
\newblock \emph{Modern Information Retrieval}.

\bibitem[{{Cui} et~al.(2020){Cui}, {Wu}, {Liu}, {Zhang}, and
  {Zhou}}]{mutual2020}
{Cui}, L.; {Wu}, Y.; {Liu}, S.; {Zhang}, Y.; and {Zhou}, M. 2020.
\newblock MuTual: A Dataset for Multi-Turn Dialogue Reasoning.
\newblock In \emph{ACL 2020: 58th annual meeting of the Association for
  Computational Linguistics}, 1406--1416.

\bibitem[{{Devlin} et~al.(2019){Devlin}, {Chang}, {Lee}, and
  {Toutanova}}]{devlin2018bert}
{Devlin}, J.; {Chang}, M.-W.; {Lee}, K.; and {Toutanova}, K. 2019.
\newblock BERT: Pre-training of Deep Bidirectional Transformers for Language
  Understanding.
\newblock In \emph{NAACL-HLT 2019: Annual Conference of the North American
  Chapter of the Association for Computational Linguistics}, 4171--4186.

\bibitem[{{Fang} et~al.(2019){Fang}, {Sun}, {Gan}, {Pillai}, {Wang}, and
  {Liu}}]{Fang2019}
{Fang}, Y.; {Sun}, S.; {Gan}, Z.; {Pillai}, R.; {Wang}, S.; and {Liu}, J. 2019.
\newblock Hierarchical Graph Network for Multi-hop Question Answering.
\newblock \emph{arXiv preprint arXiv:1911.03631} .

\bibitem[{{Gan} et~al.(2017){Gan}, {Pu}, {Henao}, {Li}, {He}, and
  {Carin}}]{gan2017learning}
{Gan}, Z.; {Pu}, Y.; {Henao}, R.; {Li}, C.; {He}, X.; and {Carin}, L. 2017.
\newblock Learning Generic Sentence Representations Using Convolutional Neural
  Networks.
\newblock In \emph{Proceedings of the 2017 Conference on Empirical Methods in
  Natural Language Processing}, 2390--2400.

\bibitem[{{Grosz}, {Weinstein}, and {Joshi}(1995)}]{grosz1995centering}
{Grosz}, B.~J.; {Weinstein}, S.; and {Joshi}, A.~K. 1995.
\newblock Centering: a framework for modeling the local coherence of discourse.
\newblock \emph{Computational Linguistics} 21(2): 203--225.

\bibitem[{{Gururangan} et~al.(2020){Gururangan}, {Marasović}, {Swayamdipta},
  {Lo}, {Beltagy}, {Downey}, and {Smith}}]{Gururangan2020}
{Gururangan}, S.; {Marasović}, A.; {Swayamdipta}, S.; {Lo}, K.; {Beltagy}, I.;
  {Downey}, D.; and {Smith}, N.~A. 2020.
\newblock Don’t Stop Pretraining: Adapt Language Models to Domains and Tasks.
\newblock In \emph{ACL 2020: 58th annual meeting of the Association for
  Computational Linguistics}, 8342--8360.

\bibitem[{{Hill}, {Cho}, and {Korhonen}(2016)}]{hill2016learning}
{Hill}, F.; {Cho}, K.; and {Korhonen}, A. 2016.
\newblock Learning distributed representations of sentences from unlabelled
  data.
\newblock In \emph{Proceedings of the 2016 Conference of the North American
  Chapter of the Association for Computational Linguistics: Human Language
  Technologies}, 1367--1377.

\bibitem[{{Joshi} et~al.(2020){Joshi}, {Chen}, {Liu}, {Weld}, {Zettlemoyer},
  and {Levy}}]{joshi2020spanbert}
{Joshi}, M.; {Chen}, D.; {Liu}, Y.; {Weld}, D.~S.; {Zettlemoyer}, L.; and
  {Levy}, O. 2020.
\newblock SpanBERT: Improving Pre-training by Representing and Predicting
  Spans.
\newblock \emph{Transactions of the Association for Computational Linguistics}
  8: 64--77.

\bibitem[{{Kipf} and {Welling}(2017)}]{kipf2016semi}
{Kipf}, T.~N.; and {Welling}, M. 2017.
\newblock Semi-Supervised Classification with Graph Convolutional Networks.
\newblock In \emph{ICLR 2017 : International Conference on Learning
  Representations 2017}.

\bibitem[{{Kiros} et~al.(2015){Kiros}, {Zhu}, {Salakhutdinov}, {Zemel},
  {Torralba}, {Urtasun}, and {Fidler}}]{kiros2015skip}
{Kiros}, R.; {Zhu}, Y.; {Salakhutdinov}, R.; {Zemel}, R.~S.; {Torralba}, A.;
  {Urtasun}, R.; and {Fidler}, S. 2015.
\newblock Skip-thought vectors.
\newblock In \emph{NIPS'15 Proceedings of the 28th International Conference on
  Neural Information Processing Systems - Volume 2}, 3294--3302.

\bibitem[{Klicpera, Bojchevski, and G{\"u}nnemann(2018)}]{klicpera2018predict}
Klicpera, J.; Bojchevski, A.; and G{\"u}nnemann, S. 2018.
\newblock Predict then propagate: Graph neural networks meet personalized
  pagerank.
\newblock \emph{arXiv preprint arXiv:1810.05997} .

\bibitem[{{Lan} et~al.(2020){Lan}, {Chen}, {Goodman}, {Gimpel}, {Sharma}, and
  {Soricut}}]{Lan2020}
{Lan}, Z.; {Chen}, M.; {Goodman}, S.; {Gimpel}, K.; {Sharma}, P.; and
  {Soricut}, R. 2020.
\newblock ALBERT: A Lite BERT for Self-supervised Learning of Language
  Representations.
\newblock In \emph{ICLR 2020 : Eighth International Conference on Learning
  Representations}.

\bibitem[{{Liu} et~al.(2019){Liu}, {Ott}, {Goyal}, {Du}, {Joshi}, {Chen},
  {Levy}, {Lewis}, {Zettlemoyer}, and {Stoyanov}}]{2019RoBERTa}
{Liu}, Y.; {Ott}, M.; {Goyal}, N.; {Du}, J.; {Joshi}, M.; {Chen}, D.; {Levy},
  O.; {Lewis}, M.; {Zettlemoyer}, L.; and {Stoyanov}, V. 2019.
\newblock RoBERTa: A Robustly Optimized BERT Pretraining Approach.
\newblock \emph{arXiv preprint arXiv:1907.11692} .

\bibitem[{{Lowe} et~al.(2015){Lowe}, {Pow}, {Serban}, and {Pineau}}]{2015Lowe}
{Lowe}, R.; {Pow}, N.; {Serban}, I.; and {Pineau}, J. 2015.
\newblock The Ubuntu Dialogue Corpus: A Large Dataset for Research in
  Unstructured Multi-Turn Dialogue Systems.
\newblock In \emph{Proceedings of the 16th Annual Meeting of the Special
  Interest Group on Discourse and Dialogue}, 285--294.

\bibitem[{{Lu} et~al.(2019){Lu}, {Zhang}, {Xie}, {Ling}, {Zhou}, and
  {Xu}}]{Lu2020}
{Lu}, J.; {Zhang}, C.; {Xie}, Z.; {Ling}, G.; {Zhou}, T.~C.; and {Xu}, Z. 2019.
\newblock Constructing Interpretive Spatio-Temporal Features for Multi-Turn
  Responses Selection.
\newblock In \emph{ACL 2019 : The 57th Annual Meeting of the Association for
  Computational Linguistics}, 44--50.

\bibitem[{Mihalcea and Tarau(2004)}]{mihalcea_textrank}
Mihalcea, R.; and Tarau, P. 2004.
\newblock {T}ext{R}ank: Bringing Order into Text.
\newblock In \emph{Proceedings of the 2004 Conference on Empirical Methods in
  Natural Language Processing}, 404--411. Barcelona, Spain: Association for
  Computational Linguistics.
\newblock \urlprefix\url{https://www.aclweb.org/anthology/W04-3252}.

\bibitem[{{Qiu} et~al.(2019){Qiu}, {Xiao}, {Qu}, {Zhou}, {Li}, {Zhang}, and
  {Yu}}]{Qiu2020}
{Qiu}, L.; {Xiao}, Y.; {Qu}, Y.; {Zhou}, H.; {Li}, L.; {Zhang}, W.; and {Yu},
  Y. 2019.
\newblock Dynamically Fused Graph Network for Multi-hop Reasoning.
\newblock In \emph{ACL 2019 : The 57th Annual Meeting of the Association for
  Computational Linguistics}, 6140--6150.

\bibitem[{{Qu} et~al.(2019){Qu}, {Yang}, {Qiu}, {Croft}, {Zhang}, and
  {Iyyer}}]{qu2019bert}
{Qu}, C.; {Yang}, L.; {Qiu}, M.; {Croft}, W.~B.; {Zhang}, Y.; and {Iyyer}, M.
  2019.
\newblock BERT with History Answer Embedding for Conversational Question
  Answering.
\newblock In \emph{Proceedings of the 42nd International ACM SIGIR Conference
  on Research and Development in Information Retrieval}, 1133--1136.

\bibitem[{{Radford} et~al.(2019){Radford}, {Wu}, {Child}, {Luan}, {Amodei}, and
  {Sutskever}}]{Radford2018}
{Radford}, A.; {Wu}, J.; {Child}, R.; {Luan}, D.; {Amodei}, D.; and
  {Sutskever}, I. 2019.
\newblock Language Models are Unsupervised Multitask Learners.

\bibitem[{{Seo} et~al.(2017){Seo}, {Kembhavi}, {Farhadi}, and
  {Hajishirzi}}]{seo2017bidirectional}
{Seo}, M.; {Kembhavi}, A.; {Farhadi}, A.; and {Hajishirzi}, H. 2017.
\newblock Bidirectional Attention Flow for Machine Comprehension.
\newblock In \emph{ICLR 2017 : International Conference on Learning
  Representations 2017}.

\bibitem[{{Su} et~al.(2019){Su}, {Shen}, {Zhang}, {Sun}, {Hu}, {Niu}, and
  {Zhou}}]{su2019improving}
{Su}, H.; {Shen}, X.; {Zhang}, R.; {Sun}, F.; {Hu}, P.; {Niu}, C.; and {Zhou},
  J. 2019.
\newblock Improving Multi-turn Dialogue Modelling with Utterance ReWriter.
\newblock In \emph{ACL 2019 : The 57th Annual Meeting of the Association for
  Computational Linguistics}, 22--31.

\bibitem[{{Tao} et~al.(2019{\natexlab{a}}){Tao}, wei {wu}, {Xu}, {Hu}, {Zhao},
  and {Yan}}]{Tao2020}
{Tao}, C.; wei {wu}; {Xu}, C.; {Hu}, W.; {Zhao}, D.; and {Yan}, R.
  2019{\natexlab{a}}.
\newblock One Time of Interaction May Not Be Enough: Go Deep with an
  Interaction-over-Interaction Network for Response Selection in Dialogues.
\newblock In \emph{ACL 2019 : The 57th Annual Meeting of the Association for
  Computational Linguistics}, 1--11.

\bibitem[{{Tao} et~al.(2019{\natexlab{b}}){Tao}, {Wu}, {Xu}, {Hu}, {Zhao}, and
  {Yan}}]{Tao2019}
{Tao}, C.; {Wu}, W.; {Xu}, C.; {Hu}, W.; {Zhao}, D.; and {Yan}, R.
  2019{\natexlab{b}}.
\newblock Multi-Representation Fusion Network for Multi-Turn Response Selection
  in Retrieval-Based Chatbots.
\newblock In \emph{Proceedings of the Twelfth ACM International Conference on
  Web Search and Data Mining}, 267--275.

\bibitem[{{Vaswani} et~al.(2017){Vaswani}, {Shazeer}, {Parmar}, {Uszkoreit},
  {Jones}, {Gomez}, {Kaiser}, and {Polosukhin}}]{Vaswani2017}
{Vaswani}, A.; {Shazeer}, N.; {Parmar}, N.; {Uszkoreit}, J.; {Jones}, L.;
  {Gomez}, A.~N.; {Kaiser}, L.; and {Polosukhin}, I. 2017.
\newblock Attention is All You Need.
\newblock In \emph{Proceedings of the 31st International Conference on Neural
  Information Processing Systems}, 5998--6008.

\bibitem[{{Wang} et~al.(2017){Wang}, {Yang}, {Wei}, {Chang}, and
  {Zhou}}]{wang2017gated}
{Wang}, W.; {Yang}, N.; {Wei}, F.; {Chang}, B.; and {Zhou}, M. 2017.
\newblock Gated Self-Matching Networks for Reading Comprehension and Question
  Answering.
\newblock In \emph{Proceedings of the 55th Annual Meeting of the Association
  for Computational Linguistics (Volume 1: Long Papers)}, volume~1, 189--198.

\bibitem[{{Wu} et~al.(2017){Wu}, {Wu}, {Xing}, {Zhou}, and {Li}}]{Wu2017}
{Wu}, Y.; {Wu}, W.; {Xing}, C.; {Zhou}, M.; and {Li}, Z. 2017.
\newblock Sequential Matching Network: A New Architecture for Multi-turn
  Response Selection in Retrieval-Based Chatbots.
\newblock In \emph{Proceedings of the 55th Annual Meeting of the Association
  for Computational Linguistics (Volume 1: Long Papers)}, volume~1, 496--505.

\bibitem[{{Xu} et~al.(2019){Xu}, {Lai}, {Feng}, and {Wang}}]{xu2019enhancing}
{Xu}, K.; {Lai}, Y.; {Feng}, Y.; and {Wang}, Z. 2019.
\newblock Enhancing Key-Value Memory Neural Networks for Knowledge Based
  Question Answering.
\newblock In \emph{NAACL-HLT 2019: Annual Conference of the North American
  Chapter of the Association for Computational Linguistics}, 2937--2947.

\bibitem[{{Ye} et~al.(2019){Ye}, {Lin}, {Liu}, {Liu}, and {Sun}}]{Martin}
{Ye}, D.; {Lin}, Y.; {Liu}, Z.; {Liu}, Z.; and {Sun}, M. 2019.
\newblock Multi-Paragraph Reasoning with Knowledge-enhanced Graph Neural
  Network.
\newblock \emph{arXiv preprint arXiv:1911.02170} .

\bibitem[{{Yeh} and {Chen}(2019)}]{yeh2019flowdelta}
{Yeh}, Y.~T.; and {Chen}, Y.-N. 2019.
\newblock FlowDelta: Modeling Flow Information Gain in Reasoning for
  Conversational Machine Comprehension.
\newblock In \emph{Proceedings of the 2nd Workshop on Machine Reading for
  Question Answering}, 86--90.

\bibitem[{{Yu} et~al.(2018){Yu}, {Dohan}, {Luong}, {Zhao}, {Chen}, {Norouzi},
  and {Le}}]{yu2018qanet}
{Yu}, A.~W.; {Dohan}, D.; {Luong}, M.-T.; {Zhao}, R.; {Chen}, K.; {Norouzi},
  M.; and {Le}, Q.~V. 2018.
\newblock QANet: Combining Local Convolution with Global Self-Attention for
  Reading Comprehension.
\newblock In \emph{International Conference on Learning Representations}.

\bibitem[{{Zhang} et~al.(2018){Zhang}, {Dai}, {Kozareva}, {Smola}, and
  {Song}}]{2017Variational}
{Zhang}, Y.; {Dai}, H.; {Kozareva}, Z.; {Smola}, A.; and {Song}, L. 2018.
\newblock Variational Reasoning for Question Answering with Knowledge Graph.
\newblock In \emph{AAAI-18 AAAI Conference on Artificial Intelligence},
  6069--6076.

\bibitem[{{Zhou} et~al.(2018){Zhou}, {Li}, {Dong}, {Liu}, {Chen}, {Zhao}, {Yu},
  and {Wu}}]{Zhou2018}
{Zhou}, X.; {Li}, L.; {Dong}, D.; {Liu}, Y.; {Chen}, Y.; {Zhao}, W.~X.; {Yu},
  D.; and {Wu}, H. 2018.
\newblock Multi-Turn Response Selection for Chatbots with Deep Attention
  Matching Network.
\newblock In \emph{ACL 2018: 56th Annual Meeting of the Association for
  Computational Linguistics}, volume~1, 1118--1127.

\end{thebibliography}

\end{document}